\documentclass[11pt,a4paper]{article}
\usepackage{times,latexsym}
\usepackage{url}
\usepackage[T1]{fontenc}
\usepackage{comment}
\usepackage{placeins} %
\usepackage{todonotes}
\usepackage{arabtex}
\usepackage[OT6,T1]{fontenc}
\usepackage[utf8]{inputenc}
\usepackage{utf8}
\usepackage{enumitem}
\usepackage{makecell}
\usepackage{tipa}  %
\usepackage{array}

\usepackage[all]{nowidow}

\usepackage[ruled,vlined]{algorithm2e}

\usepackage{booktabs}
\usepackage{ragged2e}
\newcolumntype{L}[1]{>{\RaggedRight\arraybackslash}p{#1}}
\newcolumntype{R}[1]{>{\RaggedLeft\arraybackslash}p{#1}}

\usepackage{afterpage,refcount} %
\newcommand{\setfootnotemark}{%
  \refstepcounter{footnote}%
  \footnotemark[\value{footnote}]}

\newcommand{\armenian}{\fontencoding{OT6}\fontfamily{cmr}\selectfont}
\DeclareTextFontCommand{\textarmenian}{\armenian}

\usepackage{bigfoot} %

\usepackage{amsmath} %
\usepackage{amsfonts}
\usepackage{bm}
\usepackage{mathtools}
\usepackage{nicefrac}

\usepackage[acceptedWithA]{tacl2018v2}

\newif\ifshrinkforreview
\shrinkforreviewfalse %

  \newcommand\invisiblesubsection[1]{%
  \refstepcounter{subsection}%
  \addcontentsline{toc}{subsection}{\protect\numberline{\thesubsection}#1}%
  \subsectionmark{#1}}

\newcommand{\mb}[1]{\boldsymbol{\mathbf{#1}}}
\newcommand{\mathfunc}[1]{\textsc{#1}}

\DeclareMathOperator*{\modd}{\scriptstyle\%}

\DeclarePairedDelimiter\roundbracket{(}{)}
\DeclarePairedDelimiter\squarebracket{[}{]}
\DeclarePairedDelimiter\curlybracket{\{}{\}}
\makeatletter
\def\rbr{\@ifnextchar[{\roundbracket}{\roundbracket*}}
\def\sbr{\@ifnextchar[{\squarebracket}{\squarebracket*}}
\def\cbr{\@ifnextchar[{\curlybracket}{\curlybracket*}}
\makeatother

\usepackage{xspace,mfirstuc,tabulary}

\newcommand{\namedarg}[1]{{\footnotesize\textsc{#1}}}

\newif\iftaclinstructions
\taclinstructionsfalse %
\iftaclinstructions
\renewcommand{\confidential}{}
\renewcommand{\anonsubtext}{(No author info supplied here, for consistency with
TACL-submission anonymization requirements)}
\newcommand{\instr}
\fi

\iftaclpubformat %

\else

\fi

\newcommand{\commentfn}[1]{}

\newcommand{\canine}{\textsc{Canine}\xspace}
\newcommand{\bert}{\textsc{Bert}\xspace}
\newcommand{\mbert}{m\textsc{Bert}\xspace}

\newcommand{\para}[1]{\paragraph{#1}}

\makeatletter
\def\nomarkfootnote{\xdef\@thefnmark{}\@footnotetext}
\makeatother

\setcode{utf8}

\title{\canine: Pre-training an Efficient Tokenization-Free Encoder \\ for 
Language Representation}

\author{
 Jonathan H. Clark, Dan Garrette, Iulia Turc, John Wieting \\
 Google Research \\
  {\tt\{jhclark,dhgarrette,iuliaturc,jwieting\}@google.com} \\
}

\date{}

\begin{document}
\maketitle
\begin{abstract}

Pipelined NLP systems have largely been superseded by end-to-end neural modeling, yet nearly all commonly-used models still require an explicit tokenization step. While recent tokenization approaches based on data-derived subword lexicons are less brittle than manually engineered tokenizers, these techniques are not equally suited to all languages, and the use of any fixed vocabulary may limit a model's ability to adapt. In this paper, we present \canine, a neural encoder that operates directly on character sequences---without explicit tokenization or vocabulary---and a pre-training strategy that operates either directly on characters or optionally uses subwords as a soft inductive bias. To use its finer-grained input effectively and efficiently, \canine combines downsampling, which reduces the input sequence length, with a deep transformer stack, which encodes context. \canine outperforms a comparable \mbert model by 5.7 F1 on \textsc{TyDi QA}, a challenging multilingual benchmark, despite having fewer model parameters.

\end{abstract}

\vspace{-6mm}  %
\nomarkfootnote{\canine: \textbf{C}haracter \textbf{A}rchitecture with \textbf{N}o tokenization \textbf{I}n \textbf{N}eural \textbf{E}ncoders.}
\nomarkfootnote{Code and checkpoints are available on GitHub at \mbox{\url{http://caninemodel.page.link/code}}.}
\nomarkfootnote{Published in \textit{Transactions of the Association for Computational Linguistics (TACL)}, 2022.}

\section{Introduction}

End-to-end neural models have generally replaced the traditional NLP pipeline, and with it, the error cascades and feature engineering common to such systems, preferring instead to let the model automatically induce its own sophisticated representations.
Tokenization, however, is one of few holdovers from that era, with nearly all commonly-used models today requiring an explicit preprocessing stage to segment a raw text string into a sequence of discrete model inputs.
Broadly speaking, tokenizers are generally either
carefully constructed systems of language-specific rules, which are costly, requiring both manual feature engineering and linguistic expertise,
or data-driven algorithms such as Byte Pair Encoding \cite{bpe2016}, WordPiece \cite{Wu2016}, or SentencePiece \cite{sentencepiece2018} that split strings based on frequencies in a 
corpus, which are less brittle and easier to scale, but 
are ultimately too simplistic to properly handle the wide range of linguistic phenomena that can't be captured by mere string-splitting (\S\ref{sec:pitfalls}).

The degree of sophistication required to accurately capture the full breadth of linguistic phenomena, along with the infeasibility of writing such rules by hand across all languages and domains, suggests that
explicit tokenization itself is problematic.
In contrast, 
an end-to-end model that operates directly on raw text strings would avoid these issues, instead learning to compose individual characters into its own arbitrarily complex features, with potential benefits for both accuracy and ease of use.
While this change is conceptually very simple---one could replace the subword vocabulary in a model like \bert \cite{Devlin2018} with a vocabulary made solely of individual characters---doing so leads to two immediate problems. First, the computational complexity of a transformer \cite{vaswani2017attention}, the main components in \bert as well as other models such as GPT \citep{radford2019language,brown2020language} and T5 \citep{raffel2020exploring}, grows quadratically with the length of the input. Since standard subword models have roughly four characters per subword on average, the 4x increase in input sequence length would result is a significantly slower model. Second, simply switching to a character vocabulary yields empirically poor results (\S\ref{sec:results}).

In order to enable tokenization-free modeling that overcomes these obstacles, we present \canine.
\canine is a large language encoder with a deep transformer stack at its core.
Inputs to the model are sequences of Unicode characters.\footnote{We consider splitting on Unicode characters to be tokenization-free because it depends only on the (deterministic) process defined by the Unicode standard, and not on any models, hand-crafted rules, or other linguistic knowledge.}
To represent the full space of Unicode characters\footnote{Unicode defines 1,114,112 total \textbf{codepoints}, of which only 143,698 are assigned to characters as of Unicode 13.0. This covers 154 scripts and over 900 languages.} without a vocabulary, we employ a hashing strategy.
To avoid the slowdown from increasing the sequence length, \canine uses strided convolutions to downsample input sequences to a shorter length before the deep transformer stack.

Like \bert, we pre-train \canine on the Masked Language Model (MLM) and Next Sentence Prediction (NSP) tasks. For the MLM task, \canine offers two options:
\begin{enumerate}[topsep=4pt,itemsep=-3pt]
\item A fully character-level loss that autoregressively predicts characters in masked spans.
\item A vocabulary-based loss that predicts the identities of masked subword tokens.
Critically, this tokenization is used only for the pre-training loss; tokens are never input to the encoder, and the tokenizer and subword vocabulary can be safely discarded after pre-training.
This effectively converts the hard constraint of token boundaries found in 
other models into a soft \emph{inductive bias} in \canine. 
\end{enumerate}

\noindent In this article, we contribute:
\begin{itemize}[topsep=4pt,noitemsep,label=\small$\bullet$]
     \item the first pre-trained tokenization-free deep encoder;
     \item an efficient model architecture that directly encodes long sequences of characters with speed comparable to vanilla \bert; and
     \item a model that performs no tokenization on the input, avoiding the lossy \textit{information bottleneck} associated with most pre-processing.
\end{itemize}

\section{Motivation}

\subsection{Linguistic pitfalls of tokenization}
\label{sec:pitfalls}

\begin{table}
\ifshrinkforreview
\else
\footnotesize
\fi
\centering
\begin{tabular}{lll}
\toprule
\textit{\<ك-ت-ب>} &
k-t-b &
``write'' (root form)
 \\
\textit{\<كَتَبَ>} &
\textbf{k}a\textbf{t}a\textbf{b}a &
``he wrote''
 \\
\textit{\<كَتَّبَ>} &
\textbf{k}a\textbf{tt}a\textbf{b}a &
``he made (someone) write''
 \\
\textit{\<اِكْتَتَبَ>} &
i\textbf{k}ta\textbf{t}a\textbf{b}a &
``he signed up''
 \\
\bottomrule
\end{tabular}
\caption{\label{fig:nonconcat}Non-concatenative morphology in Arabic.\setfootnotemark\label{first}
When conjugating, letters are interleaved \textit{within} the root. The root is therefore not separable from its inflection via any contiguous split.}
\end{table}
\afterpage{
    \footnotetext[\getrefnumber{first}]{From     
    \scriptsize
    \url{en.wikipedia.org/wiki/Arabic\_verbs}
    }
}

Subword tokenizers are the de-facto standard in modern NLP
\ifshrinkforreview
\else
\citep{Devlin2018,raffel2020exploring,brown2020language}.
\fi
These algorithms are limited to only simple word-splitting operations. While this is perhaps a reasonable approach for a language with impoverished morphology such as English, it is much less appropriate in the face of phenomena like
\ifshrinkforreview
\else
agglutinative morphology, non-concatenative morphology, consonant mutation, vowel harmony, etc.
\fi

Even in high-resource languages, subword models still tend to struggle on challenging domains, such as informal text, which often includes typos, spelling variation,\footnote{e.g. Spanish speakers may drop accents when typing.} transliteration, or emoji \citep{Oconnor2010}. \bert, which uses WordPiece tokenization, is sensitive to corruptions of the input, both natural typos \citep{Sun2020} and adversarial manipulations \citep{pruthi2019combating}, with some of the loss attributable to corrupted strings no longer being covered by the vocabulary.

Seemingly safe heuristics used by these algorithms, such as splitting on whitespace and punctuation, are problematic when applied to languages that do not use spaces between words
(Thai, Chinese)
or use punctuation as letters
(Hawaiian,\footnote{Hawaiian uses an apostrophe to indicate a glottal stop.} Twi\footnote{Informal Twi uses a right paren ) to represent the letter \textopeno.}). While SentencePiece does offer the option to skip whitespace splitting, it is not typically used due to poor empirical
\ifshrinkforreview
\else
performance.
\fi

Fixed vocabulary methods can also force modelers to choose between difficult preprocessing tradeoffs: should one keep accents, casing, etc. and avoid destructive preprocessing?---Or keep such orthographic information and risk important words dropping out of the frequency-based vocabulary altogether due to the presence of multiple variants of otherwise-similar words? For instance, \mbert initially removed all diacritics, thus dropping tense information in Spanish\footnote{Spanish past tense uses an accented final vowel.} and conflating many unrelated words in Vietnamese.\footnote{Vietnamese uses diacritics to indicate tones---often the only difference among several unrelated content words.}

Finally, using a fixed vocabulary during pre-training also creates complications for downstream tasks, which are subsequently tied to the same tokenizer and vocabulary used for pre-training, even if it is not well-suited for the target domain and/or end-task. 
\citet{Boukkouri2020} showed that \bert's Wikipedia+BooksCorpus WordPiece vocabulary results in excessive segmentation when fine-tuning on medical data, diminishing the benefit of pre-training as a strategy.

\subsection{Enabling better generalization}
\label{sec-better-generalization}
\label{sec:generalization}

Much as \citet{tenney2019} showed that large encoders learn elements of the classic NLP pipeline, it seems natural to let the model discover tokenization as well. With this in mind, we seek an approach that can better generalize beyond the orthographic forms encountered during pre-training.

In terms of scientific inquiry, we would like to know whether we can build models that learn how to \textit{compose} words where appropriate, and \textit{memorize} them where memorization is needed. Large frequency-derived vocabularies partially mitigate this problem by simply memorizing more, but language inherently requires aspects of both memorization and composition.
By building a model that directly engages with these issues within the small scale of word composition, we hope to enable future work studying these problems at larger scales such as phrasal constructions.

\ifshrinkforreview
\else
Practically, generalization is hindered for vocabulary elements that are slight orthographic variations, where one is very infrequent. Hypothetically, a model may estimate a very good embedding for a common vocabulary element \textit{kitten}, but a poor embedding for the less frequent element \textit{kittens} since the model has no \textit{a priori} knowledge that they are related. Embeddings that are rarely touched during pre-training will not be updated much beyond their random initializations.
\fi

\subsection{Reducing engineering effort}
\label{sec:engineering}

Mature tokenizers often include years of hand-engineered rules around special cases such as
email addresses, URLs, and handling unknown words;\footnote{For example, should a subword containing an unknown character be a separate token, or should the unknown character be separated as its own token?} even fairly minimal modern tokenizers include initial word-splitting heuristics followed by a specific algorithm and vocabulary for further breaking these tokens into subwords.

Modern pre-trained models also have many requirements throughout their lifecycle: Between the time a model is pre-trained, fine-tuned, and served---potentially months or years apart---its weights and model implementation may be converted to be compatible with another toolkit, its fine-tuning data may be tokenized in a different way, and the natural distribution of words may be quite different. All of these things introduce ample opportunities for mismatches to arise between tokenization and the vocabulary from pre-training. Yet this same pre-training paradigm presents an advantage for character models: access to a far more (unsupervised) data to learn word composition from characters; without transfer learning, this has historically been impractical for many tasks having little supervised data.

\section{\canine}
\label{sec:canine}

\canine consists of three primary components: (1) a vocabulary-free technique for embedding text; (2) a character-level model that is efficient by means of downsampling and upsampling; and (3) an effective means of performing masked language modeling on a character-level model.

\begin{figure*}[t]
\begin{center}
\vspace{-12pt}
\ifshrinkforreview
\else
\includegraphics[width=6.8in,keepaspectratio,clip,trim={0 8.5in 0 0}]{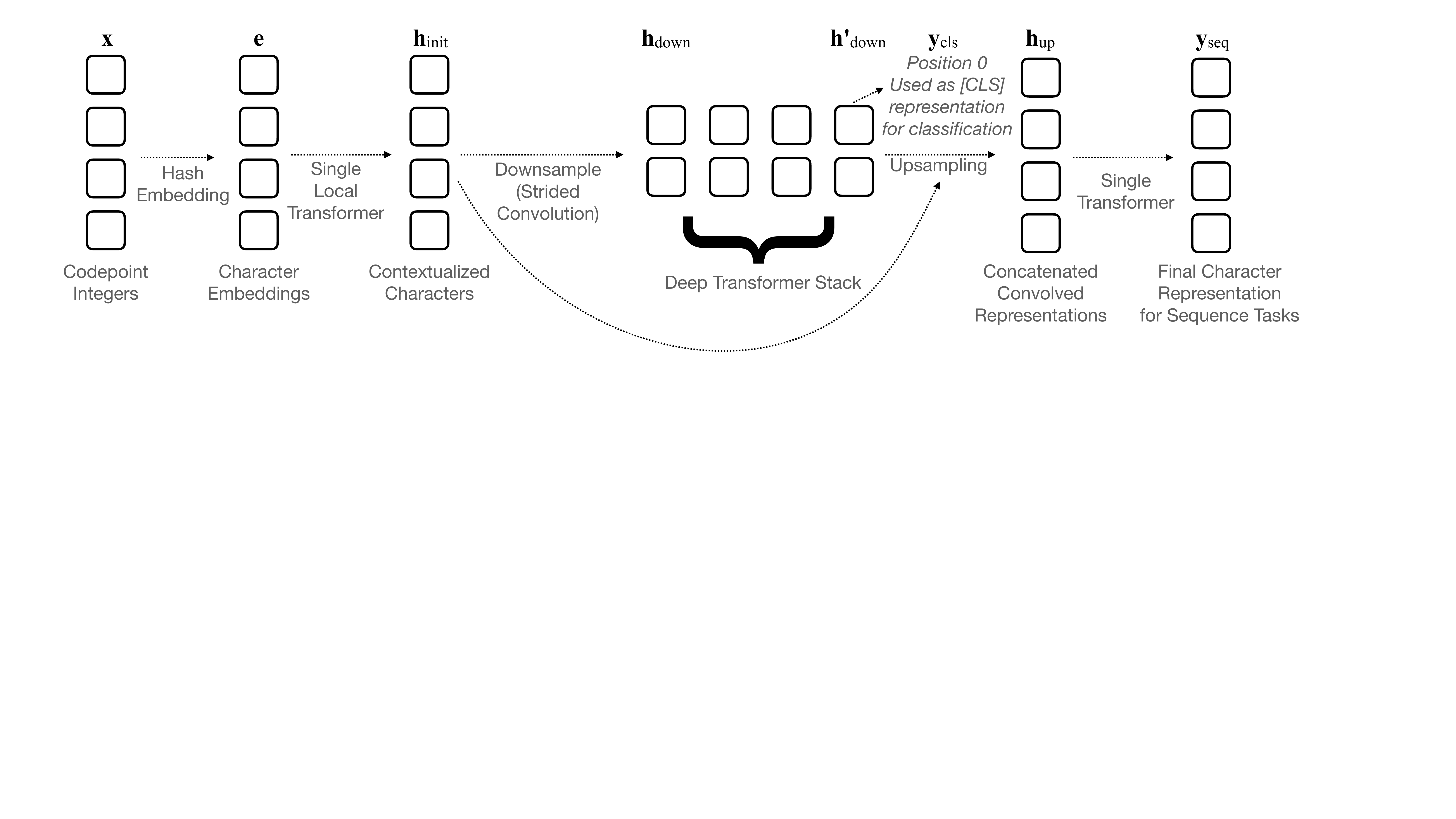}
\fi
\end{center}
\vspace{-16pt}
\caption{\label{fig:canine} \canine neural architecture.}
\end{figure*}

\subsection{Model}
\label{sec:model}

\canine is designed to be a minimally modified variant of the deep transformer stack found in modern encoders such as GPT, (m)\bert, XLM, and \mbox{XLM-R} such that its architecture is easily adoptable by other models in this family. The simplest implementation of such a character model would be to feed characters at each position in place of subwords. However, this approach would result in far more sequence positions given the same input text, leading to linearly more compute in feed forward layers and quadratically more compute in self-attention layers.

The overall form of the \canine model is the composition of a downsampling function \mathfunc{Down}, a primary encoder \mathfunc{Encode}, and an upsampling function \mathfunc{Up};\footnote{
\begin{samepage}
Enveloping the attention stack between downsampling and upsampling layers is similar to the Funnel-Transformer \citep{Dai2020}, which operates on WordPiece. However, many of its design choices (e.g., average pooling, their residual structure) did not work well in \canine.
\end{samepage}} given an input sequence of character embeddings $\mb{e} \in \mathbb{R}^{n \times d}$ with length $n$ and dimensionality $d$:
\begin{align*}
\mb{Y}_\text{seq} \leftarrow \mathfunc{Up}\left( \mathfunc{Encode}\left( \mathfunc{Down}(\mb{e}) \right) \right)
\end{align*}
where $\mb{Y}_\text{seq} \in \mathbb{R}^{n \times d}$ is the final representation for sequence prediction tasks. Similarly, for classification tasks, the model simply uses the zeroth element of the primary encoder:
\begin{align*}
\mb{y}_\text{cls} \leftarrow \left[ \mathfunc{Encode}\left( \mathfunc{Down}(\mb{e}) \right) \right]_0
\end{align*}

\para{Preprocessing} Like existing models, the input to \canine must ultimately be represented as a sequence of integers, but because the nature of characters is well-defined and standardized by Unicode, preprocessing code that would typically be hundreds or thousands of lines can be replaced by a very simple procedure: just iterate over the characters in the input string, and return their codepoint integer values (e.g., a single line of code\footnote{Python preprocessing: \texttt{[ord(c) for c in text]}} in Python). Furthermore, because codepoint values are part of the Unicode Standard, they are documented publicly, already supported by programming languages, and will not change over time, unlike arbitrary vocabulary-based IDs.

\label{sec:embed}
\para{Character hash embeddings}
\canine uses hashing \cite{Svenstrup2017} to support embedding the full space of Unicode codepoints with a relatively small number of parameters, but to reduce the chance that different codepoints will share exactly the same representation, we define a generalization of the standard hashing approach in which we apply multiple hash functions to each codepoint and concatenate the representations associated with the various hash values.

More formally, given a single codepoint\footnote{Conceptually, a \textbf{codepoint} is a character; however, a Unicode codepoint is defined precisely and unambiguously.} $x_i \in \mathbb{N}$,
we apply $K$ hash functions $\mathcal{H}_k : \mathbb{N} \rightarrow \mathbb{N}$, and look up each hashing result in its own embedding matrix\footnote{\canine uses learned embeddings, not random embedding as in other hash embeddings \cite{kaliamoorthi-etal-2019-prado}.} $\mathcal{E}_k \in \mathbb{R}^{B \times d'}$, yielding $K$ embeddings of size $d' = \nicefrac{d}{K}$, which are then concatenated into a single representation of size $d$:
\begin{align*}
e_i \leftarrow \bigoplus_k^K \mathfunc{Lookup}\left(\mathcal{H}_k(x_i) \modd B, 
\hspace{2pt}
\mathcal{E}_k\right)
\end{align*}
where $\oplus$ denotes vector concatenation.
We refer to these as the character embeddings $\mb{e} \in \mathbb{R}^{n \times d}$. In our experiments, we use $d=768$, $K=8$, and $B=16$k.\footnote{The memory footprint of these hash embeddings is equivalent to a vocabulary embedding with 16k items.}

While each individual hash function is subject to hash collisions,\footnote{This is \textit{not} a probing/chaining hash table, but rather as an \textit{approximate map}, where we expect and tolerate collisions, similar to a Bloom Map \cite{talbot2008bloom}.} 
the overall effect is minimal since each function only accounts for a small portion of the codepoint's overall embedding, and it is highly improbable that the other hash functions will produce the same collisions.

Because the model always supports all codepoints, it is possible to learn representations during fine-tuning for characters (and, by extension, words, scripts, etc.) that were never seen during pre-training, while still making use of what pre-training learned about word composition and sentence structure.

\para{Optional vocabulary-free n-grams} We can also redefine the embeddings $e_i$ above to include character n-grams, again without a fixed vocabulary, such that each n-gram order contributes equally to a summed embedding:\footnote{We use $B=15$k and $N=4$ for our n-grams.}
\begin{align*}
e_i^N \leftarrow \bigoplus_k^K \sum_j^N \mathfunc{Lookup}\left(\mathcal{H}'_k(x_{i \ldots j}) \modd B, 
\hspace{2pt}
\mathcal{E}_{j,k}\right)
\end{align*}
\vspace{-12pt}
\begin{align*}
\mathcal{H}'_k(x_{i \ldots j}) = 
\begin{cases}
    \mathcal{H}_k(x_i) \hspace{80pt} \text{if $i = j$} \\
    \mathcal{H}'_k\left(x_i + \mathcal{H}'_k(x_{(i+1) \ldots j}) \right) \hspace{6pt} \text{o/w}
\end{cases}
\end{align*}
This formulation still admits tokenization-free modeling, but provides the model with an inductive bias that favors slightly more memorization via a compute-cheap means of adding parameters. Notably, it also allows the model's input signature to remain a simple sequence of codepoints.

\paragraph{Downsampling} To make \canine efficient, we use a multi-part downsampling strategy. First, we encode characters using a single-layer 
block-wise local attention transformer. This model performs self-attention only within each block of a pre-defined size,\footnote{We use blocks of 128 characters in our experiments.} saving the quadratic cost of attention while leveraging the linguistic intuition that word composition---i.e., the kind of composition relevant in the lowest layers of the model \cite{tenney2019}---tends to happen at a very local level. Next, we use a strided convolution to reduce the number of sequence positions to be similar to that of a word piece model.\footnote{In our experiments, we found a downsampling rate of 4X to result in high quality with a speed comparable to \bert.}
Given character embeddings $\mb{e} \in \mathbb{R}^{n \times d}$ with a sequence length of $n$ characters and dimensionality $d$, we use a convolution with a stride of $r$ to downsample the sequence:
\begin{align*}
\mb{h}_\text{init} &\leftarrow \mathfunc{LocalTransformer}_1\rbr{ \mb{e} } \\
\mb{h}_\text{down} &\leftarrow \mathfunc{StridedConv}\rbr{ \mb{h}_\text{init}, \hspace{2pt} r }
\end{align*}
We refer to this output as the \textit{downsampled positions}: $\mb{h}_\text{down} \in \mathbb{R}^{m \times d}$ where $m = \nicefrac{n}{r}$ is the number of downsampled positions. In our experiments, we use $r=4$ and $n=2048$ such that $m=512$, giving \canine's primary encoder---the transformer stack---the same length as in \mbert.

\paragraph{Deep transformer stack}  
After downsampling, \canine applies a deep transformer stack with $L$ layers to the resulting downsampled positions. This is the same as the core of \bert and derivative models, and remains the core of \canine in that it accounts for the vast majority of its compute and parameters, though we note that this middle portion of the model could easily be replaced with any other sequence-to-sequence model including those with better compute performance such as Performer \cite{Choromanski2021}, Big Bird \cite{Zaheer2020}, RFA \citep{peng2021random}, ETC \cite{Ainslie2020},  etc. This portion of the model yields a new downsampled representation $\mb{h}'_\text{down} \in \mathbb{R}^{m \times d}$:
\begin{align*}
\mb{h}'_\text{down} &\leftarrow \mathfunc{Transformer}_L\rbr{ \mb{h}_\text{down} } \\
\mb{y}_\text{cls} &= [\mb{h}'_\text{down}]_0
\end{align*}
We used $L=12$ to match \mbert.

\paragraph{Upsampling} While the above architecture is sufficient for classification tasks, sequence prediction tasks require that the model expose an output layer with the same sequence length as the input (i.e., characters are the model's input and output ``API'' for tasks like tagging and span prediction).

We reconstruct a character-wise output representation by first concatenating the output of the original character transformer (above) with the downsampled representation produced by the deep transformer stack. (Note that since each downsampled position is associated with exactly $r$ characters for a downsampling rate of $r$, each position of downsampled representation is replicated $r$ times before concatenation.) More formally,
\begin{align*}
\mb{h}_\text{up} &\leftarrow \mathfunc{Conv}\left( \mb{h}_\text{init} \oplus \mb{h}'_\text{down}, w \right) \\
\mb{y}_\text{seq} &\leftarrow \mathfunc{Transformer}_1\rbr{\mb{h}_\text{up}}
\end{align*}
where $\oplus$ indicates vector concatenation of the representations (i.e. not sequences) such that \mathfunc{Conv} projects from $\mathbb{R}^{n \times 2d}$ back to $\mathbb{R}^{n \times d}$ across a window of $w$ characters.\footnote{We use $w=4$ in our experiments.} Applying a final transformer layer (standard, not local) yields a final sequence representation $\mb{y}_\text{seq} \in \mathbb{R}^{n \times d}$.

\para{Residual connections}
While the initial character encoder (before downsampling) and final character encoder (after upsampling) both represent character \textit{positions}, they conceptually have very different purposes in the network. Intuitively, we think of the initial character encoder as composing characters to create a more word-like representation, while the final character encoder is extracting the in-context representation that's relevant for predicting the ``meaning'' of the content at each position; \canine must be able to deal with additional ambiguity during upsampling since a single downsampled position may span more than one conceptual word. Because of the different roles of these induced features, we do \textit{not} use residual connections from $\mb{h}_\text{init}$ to $\mb{h}_\text{up}$.

\subsection{Pre-training}
\label{sec:pretrain}

\begin{figure*}[t]
\begin{center}
\vspace{-12pt}
\includegraphics[width=5.8in,keepaspectratio,clip,trim={0 9.5in 7.0in 0}]{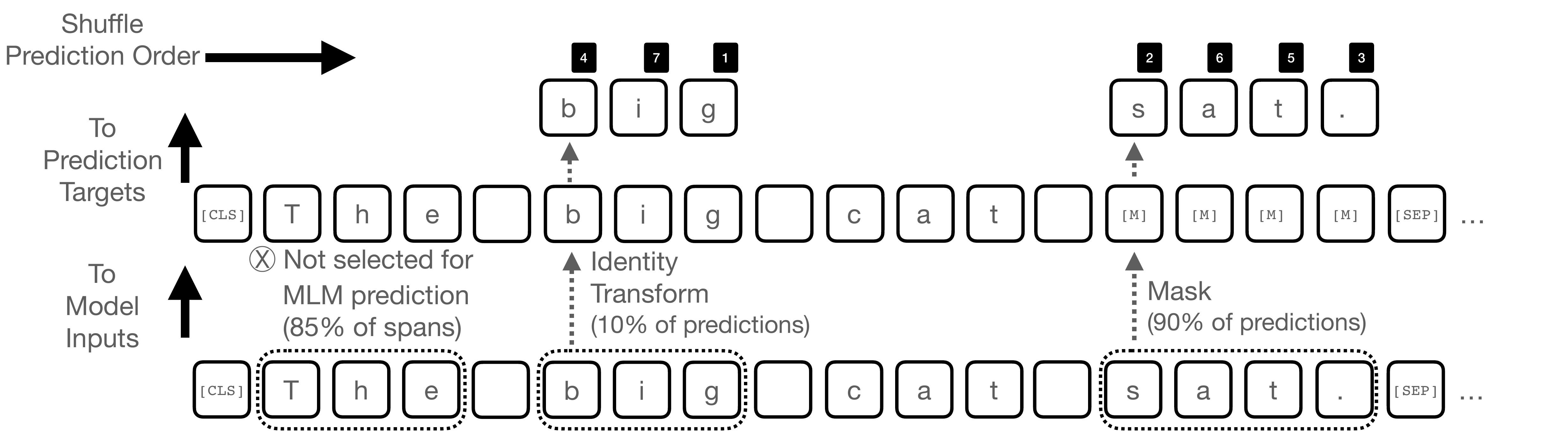}
\end{center}
\vspace{-15pt}
\caption{\label{fig:autoregressive} \canine-C pre-training data preparation (\S\ref{sec:autoregressive}). Character-wise predictions are made by an auto-regressive transformer layer that predicts then reveals one character at a time, in a shuffled order.}
\end{figure*}

Recent pre-trained models ranging from \bert to T5 have largely used variations on a masked language model (MLM) task (also known as \emph{span corruption}) as an unsupervised pre-training loss function---a means of generating synthetic examples that are not from any realistic task, yet prepare a model to learn realistic tasks in future phases of training (i.e. fine-tuning). The \canine pre-training procedure retains the MLM task, and offers two distinct strategies for computing the MLM loss---autoregressive character prediction vs. subword prediction---both of which yield a fully tokenization-free model following pre-training. In our experiments, we use only one of these losses at a time.

\subsubsection{Autoregressive Character Loss}
\label{sec:autoregressive}

\para{Span-wise masking} \canine-C is an autoregressive \underline{\textbf{c}}haracter loss that masks character spans within each sequence. These spans are chosen based on whitespace boundaries. No punctuation splitting nor other heuristics are used. All characters within the masked span are replaced by a special mask codepoint in the input.\footnote{We use codepoints in Unicode's Private Use Area block such that the input remains a valid Unicode string.}
No random subword replacement is performed as there is no subword vocabulary.\footnote{Though we expect that future work on vocabulary-free random replacement may improve quality.}

\para{Span prediction} \canine-C auto-regressively predicts the masked characters. The order of the masked positions is shuffled such that masked context is not necessarily revealed left-to-right, but rather a single character at a time. The pre-training data preparation is shown in Figure~\ref{fig:autoregressive}. Masked inputs are fed to the model as $\mathbf{x}$. The output of the \canine model $\mathbf{y}_\text{seq}$ and the embeddings $\mathbf{e}_\text{g}$ of the gold characters $\mathbf{g}$ (i.e. the character positions selected for MLM prediction) are concatenated and then fed through a small feed-forward neural network to project back to the original dimensionality $d$; these are finally shuffled and used by a single layer auto-regressive transformer with a left-to-right self-attention mask:\footnote{The left-to-right self-attention masking is with regard to the \textit{shuffled} sequence.}
\begin{align*}
\hat{\mathbf{y}} &\leftarrow \textsc{Transformer}_\textsc{AutoReg} \left( \mathbf{e}_\text{g} \oplus \mathbf{y}_\text{seq} \right)
\end{align*}
This representation $\hat{\mathbf{y}}$ is then used to predict each character. To avoid wasting time on a large output weight matrix and softmax, the gold target classes $\mb{t}$ are bucketed codepoint IDs such that $t_i = g_i \modd B$. This is similar to the strategy used in the character hash embedder (\S\ref{sec:embed}). The occassional collisions among characters is less problematic due (a) the fact that this is an encoder-only model and (b) that the embeddings must still retain contextual information in order to correctly predict characters. Because we're only predicting a relatively small subsequence of the input (15\% in our experiments), the cost of this layer is small.

\subsubsection{Subword Loss}

We also experiment with \canine-S, a \underline{\textbf{s}}ubword-based loss function, to demonstrate how a token-aware pre-training loss can still be paired with a tokenization-free model such that the tokenizer and vocabulary are discarded after pre-training.

\para{Span-wise masking} 
Like \mbert's MLM setup, each span in \canine-S  corresponds to a single subword. As with the autoregressive loss, all characters within the masked span are replaced with a special ``mask'' codepoint. Random replacements of subwords are chosen from the vocabulary of same-length subwords such that the length of the character sequence remains unchanged; more formally, given a subword selected for random replacement $x$ and a vocabulary of subwords $V$, $x$'s replacement will be drawn from the subset of $v \in V$ where $\textsc{Len}(v) = \textsc{Len}(x)$.

\para{Span prediction} Within each masked character span, \canine-S randomly selects a character position where the model will make a prediction; the model predicts the identity of the masked subword via softmax. The associated subword embeddings are discarded after pre-training.

\subsubsection{Targeted Upsampling}

By design, each final character representation (after upsampling) is a function of the output of the initial character encoder (before downsampling) and the output of the deep transformer stack---there are no inter-position dependencies across the upsampled sequence. This depends on the upsampler using position-wise feed-forward projections and a single transformer layer. During pre-training, we leverage this design to improve speed by only performing upsampling on the sequence positions that will be used by the MLM task $\mathbf{p}$.
More formally, we use the following equivalent\footnote{This highly-effective targeted upsampling optimization is the primary reason that \canine uses a full Transformer layer for the final full-length character sequence rather than a local transformer. Because a block-wise local transformer assumes uniform position-wise locality over attention blocks, it is not trivial to combine these two optimizations; the local self-attention mask would no longer be a simple block diagonal. However, this final upsampling layer is discarded for classification tasks and so does not contribute any cost. Hence, while it is possible to combine local attention and targeted upsampling, this is left as future work.} form of the $\mathfunc{Up}$ function during pre-training:
\begin{align*}
\mb{h}^*_\text{up} &\leftarrow \mathfunc{Gather}\left(\mathbf{p}, \hspace{4pt} \mb{h}_\text{up} \right) \\
\mb{y}^*_\text{seq} &\leftarrow \mathfunc{Transformer}_1\rbr{
    \namedarg{Q}=\mb{h}^*_\text{up}, \namedarg{KV}=\mb{h}_\text{up}}                  
\end{align*}

\subsubsection{Modularity}

Unlike previous models, \canine removes both the vocabulary and tokenization algorithm as fossilized parts of the final model that must be
replicated during fine-tuning and prediction. Regardless of which pre-training loss is chosen (characters or subwords), the use of these components in \canine is limited to a detail of the pre-training procedure---an \textit{inductive bias} of the loss function---that is then discarded. The fine-tuning and prediction phases of the model lifecycle never have any knowledge of what vocabulary or tokenization algorithm (if any) were used in pre-training. This allows the model to natively process untokenized data, or even process data that has been pre-processed by different tokenizers, a situation that would otherwise
introduce a significant skew between training phases.

\section{Experiments}
\label{sec:experiments}

\begin{table*}[t]
\begin{center}
\ifshrinkforreview
\else
\footnotesize
\fi
\begin{tabular}{llllR{0.35in}R{0.5in}p{0.4in}p{0.6in}p{0.6in}}
\toprule
 &  &  &  &  & \bf Examples &  & \bf \textsc{TyDiQA} & \bf \textsc{TyDiQA} \\
\bf Model & \bf Input & \bf MLM & \bf \textit{r} & \bf Length & \bf / sec & \bf Params & \bf \textsc{SelectP} & \bf \textsc{MinSpan} \\
\midrule
\mbert (public) & Subwords & Subwords & -- & 512  & -- & 179M      & 63.1 & 50.5 \\
\mbert (ours) & Subwords & Subwords & -- & 512    & 9000 & 179M    & 63.2 & 51.3 \\
& Chars & Single Chars & 1 & 2048                & 925 & 127M     & 59.5 ~\scriptsize (-3.7) & 43.7 ~\scriptsize (-7.5) \\
& Chars & Subwords & 1 & 2048               & 900 & 127M     & 63.8 ~\scriptsize (+0.6) & 50.2 ~\scriptsize (-1.0) \\

\canine-S & Chars & Subwords & 4 & 2048         & 6400 & 127M    & 66.0 ~\scriptsize (+2.8) & 52.5 ~\scriptsize (+1.2) \\

\canine-C & Chars & Autoreg. Chars & 4 & 2048         & 6050 & 127M    & 65.7 ~\scriptsize (+2.5) & 53.0
~\scriptsize (+1.7) \\

\canine-C + n-grams & Chars & Autoreg. Chars & 4 & 2048         & 5600 & 167M    & \textbf{68.1} ~\scriptsize (+4.9) & \textbf{57.0}
~\scriptsize (+5.7) \\
\bottomrule
\end{tabular}
\end{center}
\caption{\label{tab:vsbert} Direct comparison between \mbert (rows 1--2) and \canine (rows 5--7) on \textsc{TyDi QA}. Public \mbert results are taken from the \textsc{TyDi QA} paper. Rows 3 and 4 show simple baselines that yield inefficient / low-quality performance. Despite operating on 4x more sequence positions, \canine remains comparable to \mbert in terms of speed.
Pre-training example/sec are shown for our reported hardware (see Setup, \S\ref{sec:setup}).
$r$ represents the ratio for downsampling.
Parameters are calculated at fine-tuning time.
All results are averaged over 3 fine-tuning replicas. \textsc{TyDi QA} scores are F1 scores, macro-averaged across languages.
Deltas from our \mbert (the most comparable baseline) are shown in parentheses.
}
\end{table*}

\invisiblesubsection{Experimental Setup}
\label{sec:setup}

\subsubsection{Information-Seeking QA Data}

\para{\textsc{TyDi QA}: Primary Tasks}
\textsc{TyDi QA} is a dataset of information-seeking questions in 11 typologically diverse languages \cite{Clark2020}. Questions are written before answers, leading to less lexical and morphological overlap between questions and answers, which are drawn from Wikipedia. We evaluate on the primary tasks.\footnote{As opposed to the simplified \textsc{TyDiQA-GoldP} task, which is part of the \textsc{Xtreme} meta-benchmark.}

\para{Passage Selection Task (\textsc{SelectP})} Given a list of the passages in a Wikipedia article, return either the index of the passage that answers the question, or return \texttt{NULL} if the article contains no acceptable answer.

\para{Minimal Answer Span Task (\textsc{MinSpan})} Given a full Wikipedia article, return the start and end byte indices of the minimal span that completely answers the question. Alternatively, a system may indicate that the article does not contain an answer, or return \texttt{YES} or \texttt{NO} for yes/no type questions.

\subsubsection{Named Entity Recognition Data}

We also consider the task of named entity recognition (NER), which requires the model to identify which spans of a sentence correspond to entities and label the entity type. In all of our experiments, we framed the task as sequence labeling, predicting BIO-encoded span labels.

\para{CoNLL NER} We use Spanish and Dutch data from the CoNLL 2002 NER task \cite{tjong-kim-sang-2002-introduction} and English and German from the CoNLL 2003 NER task \cite{tjong-kim-sang-de-meulder-2003-introduction}, all from the newswire domain.

\para{MasakhaNER} To widen the scope of our experiments beyond Europoean languages, we also include MasakhaNER \cite{adelani2021masakhaner}, which includes ten African languages (Amharic, Hausa, Igbo, Kinyarwanda, Luganda, Luo, Nigerian Pidgin, Swahili, Wolof, and Yorùbá) with human annotations on local news text.

\subsubsection{Model Configuration}

\para{Direct comparison with \mbert}
\label{sec:vsbert}
In order to determine which pre-training architecture produces better quality downstream predictions, we compare \canine to \mbert, which we re-implemented and re-trained in order to hold as many variables as possible constant. Note that we intentionally do \textit{not} compare against public pre-trained checkpoints that use different pre-training corpora since (a) this would be a major confounding variable and (b) most publicly available pre-trained models are simply instantiations of \bert, including \mbox{XLM-R}\footnote{\mbox{XLM-R} instantiates \bert with a larger pre-training corpus, larger model size, and larger vocabulary size.} and X-STILTS.\footnote{X-STILTS performs English fine-tuning on an existing \mbox{XLM-R} checkpoint. \cite{Phang2020}}

\para{Setup} We pre-train on the multilingual Wikipedia data of \mbert, which includes 104 languages. Similarly, we reuse \mbert's exponential smoothing technique to weight the languages within the pre-training samples. We train for 124k steps with batch size 4096 (2.5 passes over the data) using the LAMB optimizer \citep{You2019} with a linearly decayed learning rate of 0.018 where 2.5\% of the steps are used for warm-up. We use a sequence length of 512 for \mbert, and 2048 for \canine, which results in 512 downsampled positions in its core deep transformer stack. We pre-train on 64 Cloud TPUs~v3\footnote{v3 TPUs have 16 GiB memory / core (128 GiB total).} for approximately one day (see results for precise timings). For both \mbert and \canine-S (\canine with the subword loss), we select 15\% of subwords for the MLM loss and predict up to 80 output positions; 80\% of these are masked in the input, 10\% are randomly replaced, and 10\% are unmodified. For \canine-C (\canine with the autoregressive character loss), we select 15\% of contiguous spans for the MLM loss and predict up to 320 output characters, and no random replacement is performed. For \textsc{TyDi QA}, we use a maximum answer length of 100 characters, which is approximately the 99$^\text{th}$ percentile answer length. Sequences longer than the maximum sequence length are zero-padded, following BERT.\footnote{Each pre-training uses approximately 24 hours on 64 TPUs (1.5k TPU-hours), so the 18 pre-trainings in Tables 2/3/4 required about 28k TPU-hours. The 18 TyDi QA experiments in these tables, each take about 1 hour on 16 TPUs, each with 3 replicas (48 TPU-hours), about 1k TPU-hours total. The 3 NER experiments in Table 5 each took 3 hours on 4 TPUs with 3 replicas each (36 TPU-hours), 108 TPU-hours total. Thus replicating the experiments in this paper would take approximately 29k TPU-hours.}

\subsection{\textsc{TyDi QA} Results}
\label{sec:results}

Our main result is shown in Table~\ref{tab:vsbert}. \mbox{\canine-S} (\canine with the subword loss) improves over \mbert in the \textsc{TyDi QA SelectP} task by 2.8 F1, while using about 30\% fewer parameters.  Similarly, \canine-C (\canine with the autoregressive character loss), improves over \mbert by 2.5 F1. Adding vocab-free character n-grams leads to even further gains over \mbert (+3.8 F1) and even more on the \textsc{MinSpan} task (+6.9 F1). A language-wise breakdown is provided in Table~\ref{tab:tydilanguages} in the appendix.

\begin{table*}[t]
\begin{center}
\scriptsize
\begin{tabular}{p{2.0in}p{4.0in}}
\toprule
\bf Question & \bf Passage Answer \\
\midrule
Chelsea ina \textbf{milikiwa} na nani? & Kwa kawaida Chelsea huvaa jezi ya blu, kaptula blu na soksi nyeupe. Nembo ya klabu imebadilishwa mara nyingi kulingana na wakati na kuboresha muonekano wa klabu. Nembo ya sasa inaonesha picha ya simba akiwa amebeba mkuki. Tangu Julai 2003, Chelsea imekuwa iki\textbf{milikiwa} na Bilionea wa Kirusi, Roman Abramovich. \\
\textit{Who owns Chelsea?} & \textit{Chelsea usually wear blue jerseys, blue shorts and white socks. The club logo has been changed many times over time and improved the club's appearance. The current emblem shows a picture of a lion carrying a spear. Since July 2003, Chelsea has been owned by Russian billionaire Roman Abramovich.} \\
\midrule
Kampuni \textbf{isambazayo} umeme nchini Kenya inaitwaje? & Kenya Power and Lighting (KPLC) ni kampuni inayohusika na maambukizi ya umeme na \textbf{usambazaji} wa umeme nchini Kenya.
\\
\textit{What is the name of the company that distributes electricity in Kenya?} & \textit{Kenya Power and Lighting (KPLC) is a company responsible for electricity transmission and distribution in Kenya.} \\
\bottomrule
\end{tabular}
\vspace{-12pt}
\end{center}
\caption{\label{tab:examples} Kiswahili examples in which \canine improved over \mbert in the \textsc{TyDi QA SelectP} task. On examining the \mbert's subword tokenization, we observe that the segmentations do not align well, putting more pressure on the model to combine them and more opportunities for some embeddings to be poorly estimated.
\textbf{Top:} The model must match a key word in the question \textit{milikiwa} (\textit{own}) to a morphological variant in the answer \textit{iki-milikiwa} (\textit{to be owned}). \mbert's WordPiece segmentation produces \textit{milik -iwa} and \textit{iki -mi -iki -wa} for these, respectively.
\textbf{Bottom:} The model must match \textit{i-sambaza-yo} (\textit{distributes}) in the question with \textit{u-sambaza-ji} (\textit{distribution}). \mbert's WordPiece segmentation produces \textit{isam -ba -za -yo} and \textit{usa -mba -zaj -i}.}
\vspace{-8pt}
\end{table*}

We also present results from some ablation models as additional baselines in rows 3-4 of Table~\ref{tab:vsbert}. First, for row~3, we simply replace \bert's subword vocabulary with a pure character vocabulary, which makes characters both the input granularity and the unit of masking and prediction for the MLM task, and observe that not only is the model 10X slower than subword-based \bert, but the quality also suffers greatly. Then, for row~4, we modify that model to use subwords for masking and MLM predictions, while keeping characters as the input granularity, and we see a substantial quality improvement, though pre-training remains extremely slow. Finally, by comparing to the full \canine model in row 5, we can see that adding the downsampling strategy improves speed by 700\%, and also leads to an additional small bump in quality. We speculate that this additional quality gain comes from giving the model a better inductive bias toward more word-like units within the deep transformer stack.

\para{Analysis} \canine fares particularly well on morphologically rich languages such as Kiswahili. Table~\ref{tab:examples} shows examples where \canine outperforms \mbert on the \textsc{TyDi QA SelectP} task. In particular, we observe examples where Kiswahili's rich morphology does not hinder the matching process for \canine.

\subsection{Ablations}
\label{sec:ablations}

In Table~\ref{tab:ablations}, we consider minor modifications to the final \canine architecture, and evaluate the effect of each on the downstream quality of the model.\footnote{These ablations were carried out during initial model development, hence comparisons to a non-final model.}

\para{Attending directly to $\mb{h}'_\text{down}$} 
Instead of attending to the character-wise sequence $\mb{h}_\text{up}$, we attend to the downsampled sequence:
\begin{align*}
\mb{y}_\text{seq}^+ = \mathfunc{Transformer}_1\rbr{\namedarg{Q}=\mb{h}_\text{up}, \namedarg{KV}=\mb{h}'_\text{down}}
\end{align*}
While this change reduces the overall FLOPS of the model due to the reduced attention computation, it does not have a major effect on pre-training throughput. However, it does substantially degrade quality.

\para{Number of hash buckets} We reduce the number of hash buckets ($B$) from 16k to 8k, meaning more (partial) collisions in embedding lookups. This significantly hinders the \textsc{MinSpan} task.

\para{Character vocab} We switch from our hash-based no-vocabulary strategy to using a normal character vocabulary (which we derive from the pre-training corpus). We observe that this underperforms the hashing approach. We speculate that this might be due to skew between the pre-training corpus and the final downstream task since not all codepoints can be included in the vocabulary.

\para{Input character dimension} We reduced the embedding size of the initial character encoder (i.e. the embedding size of $\mb{h}_\text{init}$ and $\mb{e}$---not $\mb{h}_\text{up}$ nor $\mb{y}_\text{seq}$) and observe that quality falls off rapidly.

\para{No initial transformer} We remove the local transformer from $\mb{h}_\text{init}$ and similarly observed a marked reduction in quality.

\para{Increased downsampling} While more aggressive downsampling (a factor of 5X or 6X, rather than 4X) brings substantial speed gains, the passage-level quality degrades substantially and the minimal span predictions suffer even more.

\para{No position-limited MLM} When we do not use the trick of applying the final character transformer ($\mb{y}_\text{seq}$) only to the positions that will be computed by the MLM task, we observe a large reduction in speed. Since this model is theoretically equivalent in terms of operations, we show only the speed for exposition.

\begin{table}[t]
\begin{center}
\ifshrinkforreview
\else
\footnotesize
\fi
\begin{tabular}{p{1.5in}p{0.5in}p{0.5in}}
\toprule
\bf Model & \bf \textsc{SelectP} & \bf \textsc{MinSpan} \\
\midrule
\canine-C &  65.7 & 53.0 \\

No concatenation &  17.2 & 35.6 \\

+Final-to-initial resid. &  17.3 & 35.9  \\

+Final-to-downsampled resid. &  62.0 & 50.2  \\
\bottomrule
\end{tabular}
\vspace{-8pt}
\end{center}
\caption{\label{tab:ablations2} Ablations for residuals and feature concatenation on \textsc{TyDi QA}. Rows are \textit{cumulative} (each row contains all changes from the previous).
}
\vspace{-8pt}
\end{table}

We also performed ablations aimed at exploring the effect of feature concatenation and residuals; results are in Table~\ref{tab:ablations2}. Not concatenating the downsampled representation with the initial character representation when computing $\mathbf{h}_\text{up}$ causes the model to become unstable (row 2); adding a residual from $\mathbf{h}_\text{up}$ back to $\mathbf{h}_\text{init}$ does not help (row 3). However, additionally inserting a residual from $\mathbf{h}_\text{up}$ back to $\mathbf{h}'_\text{down}$ does stabilize the model (row 4) though it does not recover the original quality.

\subsection{NER Results}
\label{sec:nerresults}

Named entity recognition is a task in which memorization is often a very effective strategy. For example, if a model has \emph{London} in its vocabulary and sees it with the label \textsc{location} during training, then it simply has to retrieve this memorized association when it sees the token \emph{London} at test time. Therefore, evaluating on NER is helpful for understanding the ways in which different models emphasize memorization vs. generalization.

As shown in Table~\ref{tab:ner}, \canine-C performs significantly worse than \mbert on NER, likely due to \mbert's memorization-friendly vocabulary. However, when (tokenization-free) n-gram features are added to \canine-C, performance rebounds, showing that it is possible to cheaply boost a model's memorization ability while remaining fully tokenization-free.

A full language-wise breakdown is provided in the appendix (Table~\ref{tab:nerlanguages}). It's worth noting that part of the performance difference on MasakhaNER is due to \mbert producing \emph{no usable outputs} for Amharic. The \mbert pre-training data does not contain Amharic (or any Amharic-script text), so it has no vocabulary entries to Amharic's script (meaning that \mbert sees only sequences of \texttt{[UNK]} on Amharic inputs). However, since \canine always supports the full Unicode space, it is able to achieve 50 F1 even though it, too, had never seen Amharic text during pre-training. We take this as validation of \canine's vocabulary-free approach. It may also be evidence that \canine exhibits cross-script transfer abilities analogous to those in \mbert \cite{pires2019}.

\para{Error analysis} \canine-C tends not to label rarer lexical items that \mbert appears to have memorized.
For example, with \canine-C, \textit{JCPenney} (a relatively rare lexical item) is not recognized as an entity. \canine-C also tends to separate long entities; for example, ``\textit{State Street Bank and Trust Company}'' is labeled as two separate spans: ``\textit{State Street Bank}'' and ``\textit{Trust Company}''; and the location \textit{TAMPA BAY} is recognized only as \textit{TAMPA}.
However, adding n-grams features
appears to mostly resolve this issue.

\begin{table}[t]
\begin{center}
\ifshrinkforreview
\else
\footnotesize
\fi
\begin{tabular}{lll}
\toprule
\bf Model & \bf CoNLL & \bf MasakhaNER \\
\midrule
\mbert (ours) & 87.8 & 72.4 \\
\canine-C &  74.0 ~\scriptsize (-13.8) & 65.5 ~\scriptsize (-6.9) \\
\canine-C + n-grams &  86.7	~\scriptsize	(-1.1) & 76.8	~\scriptsize	(+4.3) \\
\bottomrule
\end{tabular}
\end{center}
\vspace{-8pt}
\caption{\label{tab:ner} F1 scores on NER tasks.}
\end{table}

\begin{table*}[th]
\begin{center}
\ifshrinkforreview
\else
\footnotesize
\fi
\begin{tabular}{p{3.0in}p{0.7in}p{0.7in}p{0.6in}}
\toprule
              & \bf Examples            & \bf \textsc{TyDi QA}  & \bf \textsc{TyDi QA} \\
\bf Condition & \bf / sec   & \bf \textsc{SelectP}       & \bf \textsc{MinSpan}  \\
\midrule
Attend to $\mb{h}'_\text{down}$ (instead of $\mb{h}_\text{up}$) & 6400    & 64.5 & 52.2 \\
8k codepoint hash buckets (instead of 16k) & 6400 & 64.1 ~\scriptsize (-0.4) & 50.5 ~\scriptsize (-1.7) \\
Character vocab (no hashing) & 6400 & 64.6 ~\scriptsize (+/-) & 51.2 ~\scriptsize (-1.0) \\
Input character dim 384 (instead of 768) & 6600 & 62.9 ~\scriptsize (-1.2) & 49.3 ~\scriptsize (-1.2) \\
Input character dim 192 (instead of 768) & 6400 & 61.7 ~\scriptsize (-2.4) & 47.3 ~\scriptsize (-3.2) \\
No initial character transformer & 6700 & 63.2 ~\scriptsize (-1.4) & 48.3 ~\scriptsize (-2.9) \\
Downsample by a factor of 5 (instead of 4) & 7000 & 62.9 ~\scriptsize (-1.7) & 49.2 ~\scriptsize (-2.0) \\

Downsample by a factor of 6 (instead of 4) & 9200 & 62.7 ~\scriptsize (-1.9) & 47.6 ~\scriptsize (-3.6) \\
Don't limit final character transformer to MLM positions & 5200 & --- & --- \\
\midrule
\canine-S & 6400    & 66.0 & 52.5 \\
\bottomrule
\end{tabular}
\vspace{-8pt}
\end{center}
\caption{\label{tab:ablations} Ablation experiments on the \canine model with \textsc{TyDi QA} F1 scores. Deltas are shown in parentheses with regard to the top-most experiment, which serves as the baseline configuration for all experiments in this table. Each result is averaged over 3 fine-tuning and evaluation replicas.}
\end{table*}

\section{Related Work}
\label{sec:related}

\subsection{Improvements to subword tokenization}

Further improvements to standard subword tokenization like Byte Pair Encoding (BPE) \cite{bpe2016}, WordPiece \cite{Wu2016}, and SentencePiece \cite{sentencepiece2018} have been proposed. Subword regularization \citep{Kudo2018} and BPE-dropout \citep{Provilkov2019} recognize that deterministic segmentation during training limits the ability to leverage morphology and word composition; instead, they sample at random one of the multiple tokenizations of the training input, made possible by the inherent ambiguity of subword vocabularies. \citet{wang2021multiview} recently expanded on this paradigm to enforce consistency of predictions over different segmentations. Unigram LM \citep{Kudo2018}, which builds its vocabulary top-down, was shown to align with morphology better than BPE on pre-trained encoders \citep{bostrom2020byte}.

Others have built hybrid models that use multiple granularities, combining characters with tokens \citep{luong2016achieving} or different subword vocabularies \citep{Zhang2020}.

\subsection{Character-level models}

Following the larger NLP trend, character-level n-gram models \cite{huang2013learning,wieting2016charagram,bojanowski2017enriching} have mostly been replaced by neural networks. While generally lagging behind their word-level counterparts, character-level features are important for morphologically rich languages, particularly in low-resource settings \citep{garrette2013learning}.

\para{For language modeling} Character language models (CLMs) have used vanilla RNN architectures to produce distributions over sequences of characters in a purely tokenization-free manner \citep{sutskever2011generating,graves2013generating,Kwang2017,Radford2017}. Hierarchical RNNs modeled the assumption that language operates on increasing layers of abstraction: \citet{Chung2017} jointly trained a sub-module to segment the character-level input into larger spans at each layer of a stacked LSTM.

Due to the consistent lag in performance behind their word-level counterparts, attention shifted from pure CLMs towards merely \emph{character-aware} models, still reliant on traditional tokenization. Some hybrid models processed the input at character level, but predicted words from a closed vocabulary \citep{Kim2016,Gerz2018}. Others reintroduced explicit tokenization on the input side, and either generated bursts of character sequences that formed an open vocabulary \citep{kawakami2017learning} or used a character-only generator as a fallback when the main closed-vocabulary word generator produced a rare or unknown token \citep{Matthews2018,Mielke2019}. Especially after the popularization of the inherently ambiguous subword vocabularies like BPE, several studies moved beyond a single input segmentation and marginalized over all possible segmentations \citep{Van2017,buckman2018neural,grave2019training}.

Coming full circle, \citet{Kawakami2019} induced a lexicon without any explicit supervision, reverting back to pure CLMs. In a revitalized effort to bring them on-par with coarser granularities, researchers leveraged external resources such as grounding in vision \citep{Kawakami2019} or multi-task learning together with supervised morphology tasks \citep{blevins2019}. 

After the transformer \citep{vaswani2017attention} replaced RNNs as the dominant architecture in NLP, character-level models followed. \citet{AlRfou2019} showed that byte-level vanilla Transformers significantly underperform their word-level counterparts. A similar finding was reported by \citet{radford2019language}. Although the gap has been reduced \citep{Choe2019}, subword transformers remain the status quo for pure language modeling.

\para{For specific tasks} In parallel with LM efforts, the neural machine translation (NMT) community sought to solve its open-vocabulary problem via character-level modeling. \citet{luong2016achieving} proposed a hybrid model that operated mainly at the word level, but consulted a character-level LSTM for unknown words; this was a practical compromise, as their character-only model took 3 months to train. \citet{Lee2017} enabled pure character NMT by shortening the input length via convolutional, pooling, and highway layers. Notably, their many-to-English model outperformed its subword counterpart and most bilingual baselines, with a 35\% increase in training time (on a single GPU) compared to a baseline BPE-to-char model. \canine has a similar motivation, but operates in the  context of pre-trained transformers; training is 7x faster compared to a char-to-char baseline (on TPU v3), and has a 28\% increase in training time over \mbert (Table \ref{tab:vsbert}).

Character information has been leveraged for many other end tasks as well, including: text classification \citep{Zhang2015,Zhang2017}, part-of-speech tagging and NER \citep{Gillick2016,Akbik2018,Pinter2019}, named entity detection \citep{yu-etal-2018-strength}, dependency parsing \cite{Vania2018}, and machine reading comprehension \citep{Kenter2018}. Character information proved particularly useful for low-resource languages \citep{xie-etal-2018-neural}, phenomena such as code-switching and transliteration \citep{Ball2018}, and rich morphology \citep{Vania2017}, previously receiving special modeling including adaptor grammars \citep{botha-blunsom-2013-adaptor}.

\para{For transfer learning} Token-based models have also been augmented with character-level information in the context of transfer learning, where encoders trained with unsupervised objectives are repurposed to solve downstream tasks. \citet{pinter-etal-2017-mimicking} addressed the out-of-vocabulary problem of static pre-trained word embeddings by training a model to map the surface of a word to its pre-trained representation, and used it on unknown words. ELMo \citep{peters2018deep}, a bidirectional LSTM model, applied character convolutions to its whitespace-separated input tokens. Character\@\bert \citep{Boukkouri2020} ported this technique to \bert, augmenting its existing WordPiece-tokenized input. Consistent with previous observations that feeding characters into a transformer stack comes with a huge computational cost while not improving over tokenization-based approaches \citep{AlRfou2019}, a \bert model fine-tuned for semantic parsing achieved gains only when characters \emph{complemented} subwords \citep{van2020character}.

\subsection{Multilingual models}
Multilingual NLP has been dominated by deep pre-trained multilingual models whose subword vocabularies are shared across languages. Such models borrow their architectures from monolingual predecessors and apply joint training in 100+ languages, either with unsupervised LM losses: \mbert, mT5 \citep{Xue2020}, or with additional translation losses: XLM \citep{Lample2019}, \mbox{XLM-R} \citep{Conneau2019}. \citet{Chung2020} extended this by forming language clusters with per-cluster vocabularies.
To accommodate languages unseen during pre-training, \citet{wang-etal-2020-extending} extended the vocabulary and continued pre-training.

\section{Conclusion}
\label{sec:conclusion}
\vspace{-4pt}
In this article, we described \canine, which is, to our knowledge, the first pre-trained deep encoder for language understanding that uses a tokenization-free, vocabulary-free model, while surpassing the quality of models built on top of heuristic tokenizers. \canine eliminates many engineering pitfalls for practitioners and opens up new research directions for the community.

\section*{Acknowledgements}

The authors wish to thank Noah Constant, Rami Al-Rfou, Kristina Toutanova, Kenton Lee, Ming-Wei Chang, and Tim Dozat for their feedback on this work. We would also like to thank Martin Njoroge and Nanjala Misiko for their consultations on the Kiswahili examples, Diana Akrong for consulting on Twi orthography, and Waleed Ammar for consulting on Arabic morphology.

\newpage

\bibliography{CANINE-edited}
\bibliographystyle{acl_natbib}

\clearpage

\appendix
\onecolumn
\section{Appendix}
\label{sec:appendix}

\begin{center}
\begin{table}[h!]
\begin{center}
\ifshrinkforreview
\else
\footnotesize
\fi
\begin{tabular}{L{0.8in}L{0.4in}L{0.6in}L{0.6in}l}
\toprule															
\textbf{Language}	&	\textbf{\mbert}	&	\textbf{\canine-S}			&	\textbf{\canine-C}			&	\textbf{\canine-C}			\\
	&		&				&				&	\textbf{~+ n-grams}			\\
\midrule															
\multicolumn{5}{c}{\bf \textsc{SelectP}}															\\
\midrule															
(English)	&	62.2	&	58.6	~\scriptsize	(-3.6)	&	61.6	~\scriptsize	(-0.6)	&	64.6	~\scriptsize	(+2.4)	\\
Arabic	&	82.3	&	82.8	~\scriptsize	(+0.5)	&	82.5	~\scriptsize	(+0.2)	&	84.3	~\scriptsize	(+2.0)	\\
Bengali	&	58.5	&	61.8	~\scriptsize	(+3.3)	&	62.5	~\scriptsize	(+4.0)	&	66.0	~\scriptsize	(+7.5)	\\
Finnish	&	60.4	&	62.2	~\scriptsize	(+1.8)	&	63.6	~\scriptsize	(+3.2)	&	66.7	~\scriptsize	(+6.3)	\\
Indonesian	&	61.3	&	63.5	~\scriptsize	(+2.2)	&	64.2	~\scriptsize	(+2.9)	&	65.9	~\scriptsize	(+4.6)	\\
Japanese	&	46.2	&	51.7	~\scriptsize	(+5.5)	&	49.7	~\scriptsize	(+3.5)	&	51.2	~\scriptsize	(+5.0)	\\
Korean	&	60.2	&	60.3	~\scriptsize	(+0.1)	&	59.7	~\scriptsize	(-0.5)	&	60.6	~\scriptsize	(+0.4)	\\
Russian	&	62.2	&	64.6	~\scriptsize	(+2.4)	&	65.6	~\scriptsize	(+3.4)	&	68.5	~\scriptsize	(+6.3)	\\
Swahili	&	58.8	&	67.8	~\scriptsize	(+9.0)	&	67.0	~\scriptsize	(+8.2)	&	67.2	~\scriptsize	(+8.4)	\\
Telugu	&	81.0	&	82.5	~\scriptsize	(+1.5)	&	81.1	~\scriptsize	(+0.1)	&	84.6	~\scriptsize	(+3.6)	\\
Thai	&	61.1	&	62.8	~\scriptsize	(+1.7)	&	61.2	~\scriptsize	(+0.1)	&	65.8	~\scriptsize	(+4.7)	\\
\textbf{Macro Avg}	&	63.2	&	66.0	~\scriptsize	(+2.8)	&	65.7	~\scriptsize	(+2.5)	&	68.1	~\scriptsize	(+4.9)	\\
\midrule															
\multicolumn{5}{c}{\bf \textsc{MinSpan}}															\\
\midrule															
(English)	&	46.0	&	46.3	~\scriptsize	(+0.3)	&	49.0	~\scriptsize	(+3.0)	&	51.8	~\scriptsize	(+5.8)	\\
Arabic	&	70.7	&	66.9	~\scriptsize	(-3.8)	&	65.6	~\scriptsize	(-5.1)	&	73.0	~\scriptsize	(+2.3)	\\
Bengali	&	47.3	&	46.7	~\scriptsize	(-0.6)	&	52.5	~\scriptsize	(+5.2)	&	57.1	~\scriptsize	(+9.8)	\\
Finnish	&	51.1	&	53.0	~\scriptsize	(+1.9)	&	53.8	~\scriptsize	(+2.7)	&	57.1	~\scriptsize	(+6.0)	\\
Indonesian	&	52.2	&	53.6	~\scriptsize	(+1.4)	&	54.4	~\scriptsize	(+2.2)	&	56.8	~\scriptsize	(+4.6)	\\
Japanese	&	36.1	&	40.3	~\scriptsize	(+4.2)	&	40.7	~\scriptsize	(+4.6)	&	42.0	~\scriptsize	(+5.9)	\\
Korean	&	36.8	&	35.7	~\scriptsize	(-1.1)	&	36.5	~\scriptsize	(-0.3)	&	39.9	~\scriptsize	(+3.1)	\\
Russian	&	45.6	&	46.7	~\scriptsize	(+1.1)	&	47.2	~\scriptsize	(+1.6)	&	51.5	~\scriptsize	(+5.9)	\\
Swahili	&	49.4	&	59.0	~\scriptsize	(+9.6)	&	57.6	~\scriptsize	(+8.2)	&	59.2	~\scriptsize	(+9.8)	\\
Telugu	&	75.6	&	75.2	~\scriptsize	(-0.4)	&	74.2	~\scriptsize	(-1.4)	&	79.7	~\scriptsize	(+4.1)	\\
Thai	&	48.4	&	47.9	~\scriptsize	(-0.5)	&	47.1	~\scriptsize	(-1.3)	&	54.2	~\scriptsize	(+5.8)	\\
\textbf{Macro Avg}	&	51.3	&	52.5	~\scriptsize	(+1.2)	&	53.0	~\scriptsize	(+1.7)	&	57.0	~\scriptsize	(+5.7)	\\
\bottomrule															\end{tabular}
\end{center}
\caption{\label{tab:tydilanguages} Language-wise breakdown for \textsc{TyDi QA} primary tasks. English is parenthesized because it is not included in the overall score calculation for \textsc{TyDi QA}.}
\end{table}
\end{center}

\begin{table*}[h!]
\begin{center}
\ifshrinkforreview
\else
\footnotesize
\fi
\begin{tabular}{lL{0.4in}L{0.7in}l}
\toprule											
\textbf{Language}	&	\textbf{\mbert}	&	\textbf{\canine-C}			&	\textbf{\canine-C}			\\
	&		&				&	\textbf{~+ n-grams}			\\
\midrule											
\multicolumn{4}{c}{\bf \textsc{CoNLL}}											\\
\midrule											
Dutch	&	90.2	&	74.7	~\scriptsize	(-15.5)	&	88.5	~\scriptsize	(-1.7)	\\
English	&	91.1	&	79.8	~\scriptsize	(-11.3)	&	89.8	~\scriptsize	(-1.3)	\\
German	&	82.5	&	64.1	~\scriptsize	(-18.4)	&	82.1	~\scriptsize	(-0.4)	\\
Spanish	&	87.6	&	77.4	~\scriptsize	(-10.2)	&	86.5	~\scriptsize	(-1.1)	\\
\textbf{Macro Avg}	&	87.8	&	74.0	~\scriptsize	(-13.8)	&	86.7	~\scriptsize	(-1.1)	\\
\midrule											
\multicolumn{4}{c}{\bf \textsc{MasakhaNER}}											\\
\midrule											
Amharic	&	0.0	&	44.6	~\scriptsize	(+44.6)	&	50.0	~\scriptsize	(+50.0)	\\
Hausa	&	89.3	&	76.1	~\scriptsize	(-13.2)	&	88.0	~\scriptsize	(-1.3)	\\
Igbo	&	84.6	&	75.6	~\scriptsize	(-9.0)	&	85.0	~\scriptsize	(+0.4)	\\
Kinyarwanda	&	73.9	&	58.3	~\scriptsize	(-15.6)	&	72.8	~\scriptsize	(-1.1)	\\
Luganda	&	80.2	&	69.4	~\scriptsize	(-10.8)	&	79.6	~\scriptsize	(-0.6)	\\
Luo	&	75.8	&	63.4	~\scriptsize	(-12.4)	&	74.2	~\scriptsize	(-1.6)	\\
Nigerian Pidgin	&	89.8	&	66.6	~\scriptsize	(-23.2)	&	88.7	~\scriptsize	(-1.1)	\\
Swahili	&	87.1	&	72.7	~\scriptsize	(-14.4)	&	83.7	~\scriptsize	(-3.4)	\\
Wolof	&	64.9	&	60.7	~\scriptsize	(-4.2)	&	66.5	~\scriptsize	(+1.6)	\\
Yorùbá	&	78.7	&	67.9	~\scriptsize	(-10.8)	&	79.1	~\scriptsize	(+0.4)	\\
\textbf{Macro Avg}	&	72.4	&	65.5	~\scriptsize	(-6.9)	&	76.8	~\scriptsize	(+4.3)	\\
\bottomrule											
\end{tabular}
\end{center}
\caption{\label{tab:nerlanguages} Language-wise breakdown for Named Entity Recognition for the CoNLL and MasakhaNER datasets (labeled F1). \mbert obtains a score of zero on Amharic due to having no vocabulary entries in the Amharic script.}
\end{table*}

\end{document}